\documentclass[letterpaper]{article} 

\usepackage{aaai2026}  
\usepackage{times}  
\usepackage{helvet}  
\usepackage{courier}  
\usepackage[hyphens]{url}  
\usepackage{graphicx} 
\usepackage{natbib}  
\usepackage{caption} 
\usepackage{algorithm}
\usepackage{algorithmic}

\usepackage{newfloat}
\usepackage{listings}
\DeclareCaptionStyle{ruled}{labelfont=normalfont,labelsep=colon,strut=off} 
\floatstyle{ruled}
\newfloat{listing}{tb}{lst}{}
\floatname{listing}{Listing}

\usepackage{soul}
\setuldepth{Qyg}     
\usepackage{amsmath}  
\usepackage{cleveref}
\usepackage{xcolor}
\usepackage{amssymb}  
\usepackage{amsfonts}  
\usepackage{booktabs}  
\usepackage{colortbl}  

\usepackage{textcomp}  
\newcommand{\midtilde}{\raisebox{.5ex}{\texttildelow}}

\title{
Predicting Video Slot Attention \ul{Q}ueries from \ul{Rand}om \ul{S}lot-\ul{F}eature Pairs
}

\author{
Rongzhen Zhao\textsuperscript{\rm 1},
Jian Li\textsuperscript{\rm 2},
Juho Kannala\textsuperscript{\rm 3,\rm 4},
Joni Pajarinen\textsuperscript{\rm 1}
}

\affiliations{
\textsuperscript{\rm 1}Department of Electrical Engineering and Automation, Aalto University, Espoo, Finland\\
\textsuperscript{\rm 2}Gaoling School of Artificial Intelligence, Renmin University of China, Beijing, China\\
\textsuperscript{\rm 3}Department of Computer Science, Aalto University, Espoo, Finland\\
\textsuperscript{\rm 4}Center for Machine Vision and Signal Analysis, University of Oulu, Oulu, Finland\\

\{rongzhen.zhao, juho.kannala, joni.pajarinen\}@aalto.fi,
lijian2022@ruc.edu.cn
}

\begin{document}

\maketitle

\begin{abstract}
Unsupervised video Object-Centric Learning (OCL) is promising as it enables object-level scene representation and understanding as we humans do.
Mainstream video OCL methods adopt a recurrent architecture: An aggregator aggregates current video frame into object features, termed slots, under some queries; A transitioner transits current slots to queries for the next frame.
This is an effective architecture but all existing implementations both (\textit{i1}) neglect to incorporate next frame features, the most informative source for query prediction, and (\textit{i2}) fail to learn transition dynamics, the knowledge essential for query prediction.
To address these issues, we propose Random Slot-Feature pair for learning Query prediction (RandSF.Q): (\textit{t1}) We design a new transitioner to incorporate both slots and features, which provides more information for query prediction; (\textit{t2}) We train the transitioner to predict queries from slot-feature pairs randomly sampled from available recurrences, which drives it to learn transition dynamics.
Experiments on scene representation demonstrate that our method surpass existing video OCL methods significantly, e.g., up to 10 points on object discovery, setting new state-of-the-art. Such superiority also benefits downstream tasks like scene understanding.
\end{abstract}

\begin{links}
\link{Source Code, Model Checkpoints, Training Logs}{https://github.com/Genera1Z/RandSF.Q}
\end{links}

\begin{figure}[t]
\centering
\hspace*{-0.4em}
\includegraphics[width=0.48\textwidth]{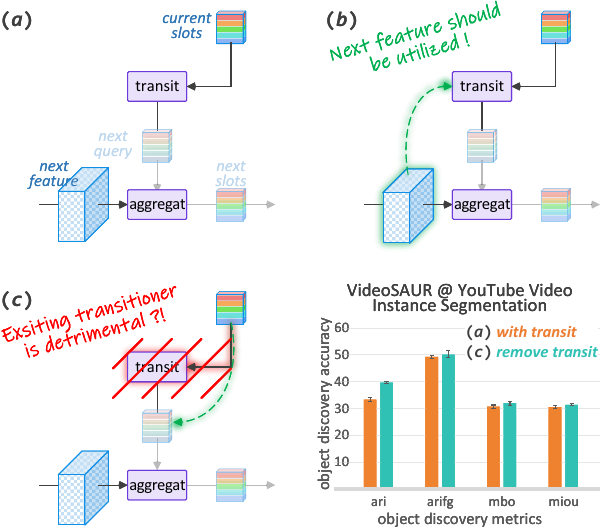}
\caption{
(\textit{a}) Mainstream video OCL adopts a recurrent architecture, where a transitioner transits current slots into the query for next video frame, and an aggregator aggregates the next frame feature into slots under the query.
(\textit{b}) Our intuitive observation: To predict next query using the transitioner, next frame feature is already available and very informative thus should also be utilized.
(\textit{c}) Our empirical observation: By removing the transitioner and using current slots directly as next query, the algorithm works even better -- Existing transitioner is not effectively learned.
}
\label{fig:teaser}
\end{figure}

\section{Introduction}
\label{sect:introduction}

By video Object-Centric Learning (OCL) \cite{singh2021slate,zadaianchuk2024videosaur}, objects in the video can be discovered and represented as respective single feature vectors, termed slots, as well as tracked across frames. This is achieved under self-supervision, by simply forcing the slots to reconstruct the input video frames in some format.
Representing a visual scene described by a video as object-level features is not only cognitively plausible as we humans perceive the world similarly \cite{palmeri2004visual,cavanagh2011visual}, but also practically feasible as a visual scene evolves upon dynamics among objects.
This is why OCL is gaining popularity in applications like scene representing \cite{locatello2020slotattent}, dynamics modeling \cite{wu2022slotformer}, planning and decision-making \cite{palmeri2004visual,ferraro2025focus} lately.

Mainstream video OCL methods \cite{singh2022steve,kipf2021savi} adopt a recurrent architecture, as shown in \Cref{fig:teaser} (\textit{a}). 
Given some query vectors, the \textit{aggregator}, a Slot Attention \cite{locatello2020slotattent} module, aggregates current video frame into slots, where each of such feature vectors represents an object. 
For next frame aggregation, the \textit{transitioner}, typically a Transformer encoder block \cite{vaswani2017transformer}, transits current slots into the next query to provide temporally consistent representations for objects in dynamically evolving scenes, ensuring coherent tracking and identity preservation across time.
Despite its overwhelmingly wide adoptation, does the query prediction via the transitioner between frames really learn the knowledge to effectively power OCL on videos?

To the best of our knowledge, all existing implementations of this recurrent architecture have not acknowledged or explored the following two issues:
\begin{itemize}
\item[\small\textit{i1}] Although visual features of the next frame are already available and are much more informative, existing transitioners predict the next query only based on current (maybe along with previous) slots without leveraging available future information, as shown in \Cref{fig:teaser} (\textit{b});
\item[\small\textit{i2}] Despite the primacy of transition dynamics knowledge in query prediction, existing transitioners lack the inductive bias to learn it, making themselves not only ineffective but even detrimental, as shown in \Cref{fig:teaser} (\textit{c}). 
\end{itemize}

\begin{figure*}[t]
\centering
\includegraphics[width=\linewidth]{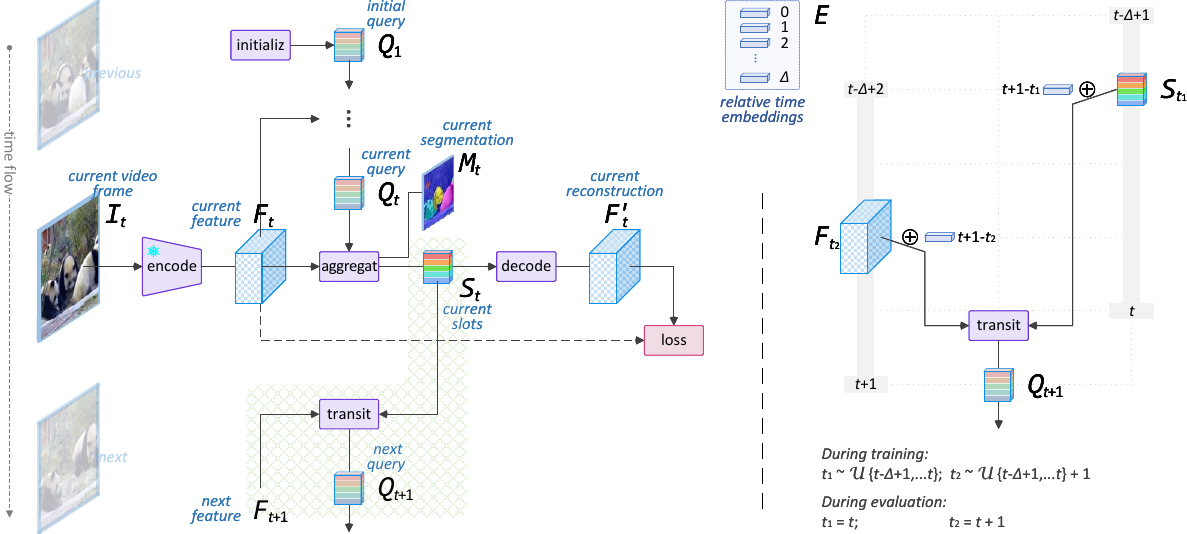}
\caption{
Our model architecture.
(\textit{left}) \textbf{Our method is built upon SlotContrast} \cite{manasyan2025slotcontrast}. A frozen DINO2 \cite{oquab2023dino2} model \textit{encodes} current video frame $\boldsymbol{I}_t$ into current feature $\boldsymbol{F}_t$; A Slot Attention \cite{locatello2020slotattent} module \textit{aggregates} $\boldsymbol{F}_t$ into current object-level vectors, slots $\boldsymbol{S}_t$, under current query $\boldsymbol{Q}_t$; A Transformer decoder block \cite{vaswani2017transformer} \textit{transits} $\boldsymbol{S}_t$ conditioned on next feature $\boldsymbol{F}_{t+1}$ to next query $\boldsymbol{Q}_{t+1}$; A random Transformer decoder \cite{zhao2025dias} \textit{decodes} $\boldsymbol{S}_t$ into current reconstruction $\boldsymbol{F}'_t$. The \textit{objective} is minimizing difference between $\boldsymbol{F}_t$ and $\boldsymbol{F}'_t$.
(\textit{right}) \textbf{How our transitioner works}. To effectively learn transition dynamics, our transitioner explores slots $\boldsymbol{S}_{t_1}$ and feature $\boldsymbol{F}_{t_2}$ at any past time point within window size $\mathit{\Delta}$ to predict next query $\boldsymbol{Q}_{t+1}$ during \textit{training}. Relative time embeddings are added to $\boldsymbol{S}_{t_1}$ and $\boldsymbol{F}_{t_2}$ to indicate their offset from $t+1$. To maximize prediction accuracy, our trainsitioner exploits only the most recent slots $\boldsymbol{S}_t$ and feature $\boldsymbol{F}_{t+1}$ to predict $\boldsymbol{Q}_{t+1}$ during \textit{evaluation}.
The left sub-figure is adapted from \cite{zhao2025vvo}.
}
\label{fig:solution}
\end{figure*}

To address the above issues, we propose RandSF.Q, which uses Random Slot-Feature pairs as transition input to enhance the learning of Query prediction. Our method tackles these limitations through two key techniques:
\begin{itemize}
\item[\small\textit{t1}] To provide more information for query prediction, we design a new transitioner, which for the first time incorporates both current slots and features of the next video frame as its input, as shown in \Cref{fig:solution} (\textit{a});
\item[\small\textit{t2}] To enforce the learning of transition dynamics, we train the transitioner to predict query for the next frame from slot-feature pairs that are randomly sampled from available recurrences, as shown in \Cref{fig:solution} (\textit{b}).
\end{itemize}

Besides two novel techniques mentioned above, our contributions also include:
\begin{itemize}
\item New state-of-the-art on video object discovery tasks, as well as consistent performance boosts on downstream tasks like object recognition and video prediction;
\item Straightforward quantification on our transitioner's query prediction capability.
\end{itemize}

\section{Related Work}
\label{sect:related_work}

In this section, we briefly review existing advances of OCL on video, compare with self-supervised object tracking, and focus on existing query prediction techniques.

\textbf{Video object-centric learning}. 
Given that Slot Attention has been central to video OCL since its inception, we classify existing methods as classical ones \cite{jiang2019scalor, kosiorek2018sqair, van2018relationalnem, veerapaneni2020op3, burgess2019monet} and modern ones \cite{kabra2021simone, aydemir2023solv, kipf2021savi, elsayed2022savipp, safadoust2023multi, singh2022steve, traub2022loci, zadaianchuk2024videosaur, zoran2021parts, qian2023smtc}. 
Modern methods' performance has improved a lot even on real-world videos under self-supervision.
Our work follows this line of research and we omit the word ``modern" for simplicity.
Video OCL is basically image OCL with recurrent module connecting between frames, i.e., current slots are transited to next query. 
Like image OCL, decoders, which are mixture-based \cite{locatello2020slotattent, seitzer2023dinosaur}, auto-regressive \cite{singh2021slate, kakogeorgiou2024spot, zhao2025dias} or denoising-based \cite{jiang2023lsd, wu2023slotdiffuz}, are also applicable to video OCL;
Decoding targets, which are quantized with different inductive biases \cite{zhao2025gdr, zhao2025msf, zhao2025vvo}, are applicable to video OCL too.
Specific to video OCL, temporal intrinsics like temporal prediction \cite{zadaianchuk2024videosaur} and consistency \cite{aydemir2023solv, manasyan2025slotcontrast} can be utilized to enhance the performance.
We focus on temporal inductive biases.

\textbf{Unsupervised video object tracking}.
Unsupervised Video Object Tracking (VOT) resembles \cite{li2019joint, yuan2020self, lai2020cvpr, wang2021unsupervised, xu2021iccv, shen2022cvpr, li2022cvpr} video OCL very much, because both are query-based and the query/slots should be maintained consistently across frames under self-supervision.
However, VOT query only covers interested objects, while video OCL query covers all objects and the background.
Essentially, unsupervised VOT only cares about temporal consistency of query representations while video OCL also demand slot representations to be informative enough to represent the scene.
Video OCL slots, the sub-symbolic representations, can be directly employed in downstream scene representation and understanding, while VOT representations can not.
Comparison with unsupervised VOT methods is of out scope.

\textbf{Query prediction}. 
It is query prediction that adapts image OCL to video OCL \cite{singh2022steve}. 
Most video OCL methods employ a single Transformer encoder block to transit current slots into next query, i.e., the naive recurrent architecture of aggreagtion-transition-aggregation... \cite{zhao2025vvo}.
Well-recognized works like STEVE \cite{singh2022steve}, SAVi \cite{kipf2021savi}, SAVi++ \cite{elsayed2022savipp}, VideoSAUR \cite{zadaianchuk2024videosaur} and SlotContrast \cite{manasyan2025slotcontrast} all take such design. 
There is no transitioner in SOLV \cite{aydemir2023solv} because global slots are aggregated for all frames, which however requires all frames to be available once together.
A few works STATM \cite{li2025statm} and SlotPi \cite{li2025slotpi} fuse architectures of video OCL \cite{kipf2021savi} and world model \cite{wu2022slotformer} and predict next query as auto-regression, which can be taken as an advanced recurrent architecture.
However, none of them acknowledge or explore issues (\textit{i1}) and (\textit{i2}), which are directly addressed by our method.

\section{Proposed Method}
\label{sect:proposed_method}

In this section, we formulate our RandSF.Q -- random slot-feature pair for effective query prediction learning and latest pair for informative prediction. 
This is realized with a novel transitioner and the corresponding novel training and evaluation strategies upon the recurrent aggregation-transition architecture of mainstream video OCL methods.

\subsection{Overall Model Architecture}

As shown in \Cref{fig:solution} (\textit{left}), we build our whole method upon SlotContrast \cite{manasyan2025slotcontrast}, the latest state-of-the-art. Our model consists of an encoder $\boldsymbol{\phi}_\mathrm{e}$, an aggregator $\boldsymbol{\phi}_\mathrm{a}$, a decoder $\boldsymbol{\phi}_\mathrm{d}$ and a transitioner $\boldsymbol{\phi}_\mathrm{t}$.

At time step $t$, the \textbf{encoder} $\boldsymbol{\phi}_\mathrm{e}$ encodes current video frame $\boldsymbol{I}_t \in \mathbb{R} ^ {h_0 \times w_0 \times c_0}$ into current feature $\boldsymbol{F}_t \in \mathbb{R} ^ {h \times w \times c}$:
\begin{equation}
\label{eq:encode}
\boldsymbol{\phi}_\mathrm{e} : \boldsymbol{I}_t \rightarrow \boldsymbol{F}_t
\end{equation}
where $\boldsymbol{\phi}_\mathrm{e}$ is parameterized as a DINO2 \cite{oquab2023dino2} model, which is a pretrained vision foundation model and is always frozen; $h_0$, $w_0$ and $c_0$ denote the input resolution, i.e., height, width and channel, while $h$, $w$ and $c$ are that after encoding.

Then the \textbf{aggregator} $\boldsymbol{\phi}_\mathrm{a}$ aggregates objects and background information scattered in the super-pixels of current feature $\boldsymbol{F}_t$ respectively into corresponding object-level feature vectors, i.e., slots $\boldsymbol{S}_t \in \mathbb{R} ^ {s \times c}$, by iteratively refining current query vectors $\boldsymbol{Q}_t \in \mathbb{R} ^ {s \times c}$ and current segmentation masks $\boldsymbol{M}_t \in \mathbb{R} ^ {h \times w}$:
\begin{equation}
\label{eq:aggregat}
\boldsymbol{\phi}_\mathrm{a} : \boldsymbol{Q}_t, \boldsymbol{F}_t \rightarrow \boldsymbol{S}_t, \boldsymbol{M}_t
\end{equation}
where $\boldsymbol{\phi}_\mathrm{a}$ is parameterized as typical Slot Attention \cite{locatello2020slotattent}; $s$ is the number of query vectors; $\boldsymbol{M}_t$ is binarized from $\boldsymbol{S}_t$'s attention maps along the slot dimension.
Because Slot Attention is a kind of attention, $\boldsymbol{Q}_t$ and $\boldsymbol{S}_t$ always have identical tensor shapes. $\boldsymbol{M}_t$ can be used for object discovery, while $\boldsymbol{S}_t$ can be used for downstream scene representation and understanding.

Besides, the \textbf{transitioner} $\boldsymbol{\phi}_\mathrm{r}$ transits current slots $\boldsymbol{S}_t$ into the query for next video frame $\boldsymbol{Q}_{t+1} \in \mathbb{R} ^ {s \times c}$ conditioned on next frame feature $\boldsymbol{F}_{t+1} \in \mathbb{R} ^ {h \times w \times c}$:
\begin{equation}
\label{eq:transit}
\boldsymbol{\phi}_\mathrm{r} : \boldsymbol{S}_t, \boldsymbol{F}_{t+1} \rightarrow \boldsymbol{Q}_{t+1}
\end{equation}
where $\boldsymbol{\phi}_\mathrm{r}$ is parameterized as a single Transformer decoder block \cite{vaswani2017transformer}. 
Aggregator $\boldsymbol{\phi}_\mathrm{a}$ and transitioner $\boldsymbol{\phi}_\mathrm{r}$ form the aggregation-transition recurrence, which powers OCL through video frames.
Please note that \textbf{extra specific processes} on $\boldsymbol{S}_t$ and $\boldsymbol{F}_{t+1}$ here are needed during training and evaluation, but are omitted for simplicity. We will detail them in the following two subsections.

Last, the \textbf{decoder} $\boldsymbol{\phi}_\mathrm{d}$ decodes current slots $\boldsymbol{S}_t$ into the reconstruction $\boldsymbol{F}'_t \in \mathbb{R} ^ {h \times w \times c}$ of current feature $\boldsymbol{F}_t$:
\begin{equation}
\label{eq:decode}
\boldsymbol{\phi}_\mathrm{d} : \boldsymbol{S}_t \rightarrow \boldsymbol{F}'_t
\end{equation}
where $\boldsymbol{\phi}_\mathrm{d}$ is parameterized as a random auto-regressive Transformer decoder \cite{zhao2025dias}. The input clue needed in such decoding is omitted for simplicity.

The \textbf{objective} is to minimize the difference between current prediction $\boldsymbol{F}'_t$ and $\boldsymbol{F}_t$:
\begin{equation}
\label{eq:objective}
\arg\min_{ \boldsymbol{\phi}_\mathrm{a}, \boldsymbol{\phi}_\mathrm{r}, \boldsymbol{\phi}_\mathrm{d} } \mathrm{MSE} ( \{ \boldsymbol{F}'_t \}_{t=1}^T , \mathrm{sg} ( \{ \boldsymbol{F}_t \}_{t=1}^T ) )
\end{equation}
where $\mathrm{MSE}(\cdot, \cdot)$ is mean squared error; $\mathrm{sg}(\cdot)$ is stopping gradient; $T$ is the total number frames in a video.
The auxiliary loss \cite{manasyan2025slotcontrast} is omitted for simplicity.

\begin{figure*}
\centering
\includegraphics[width=\textwidth]{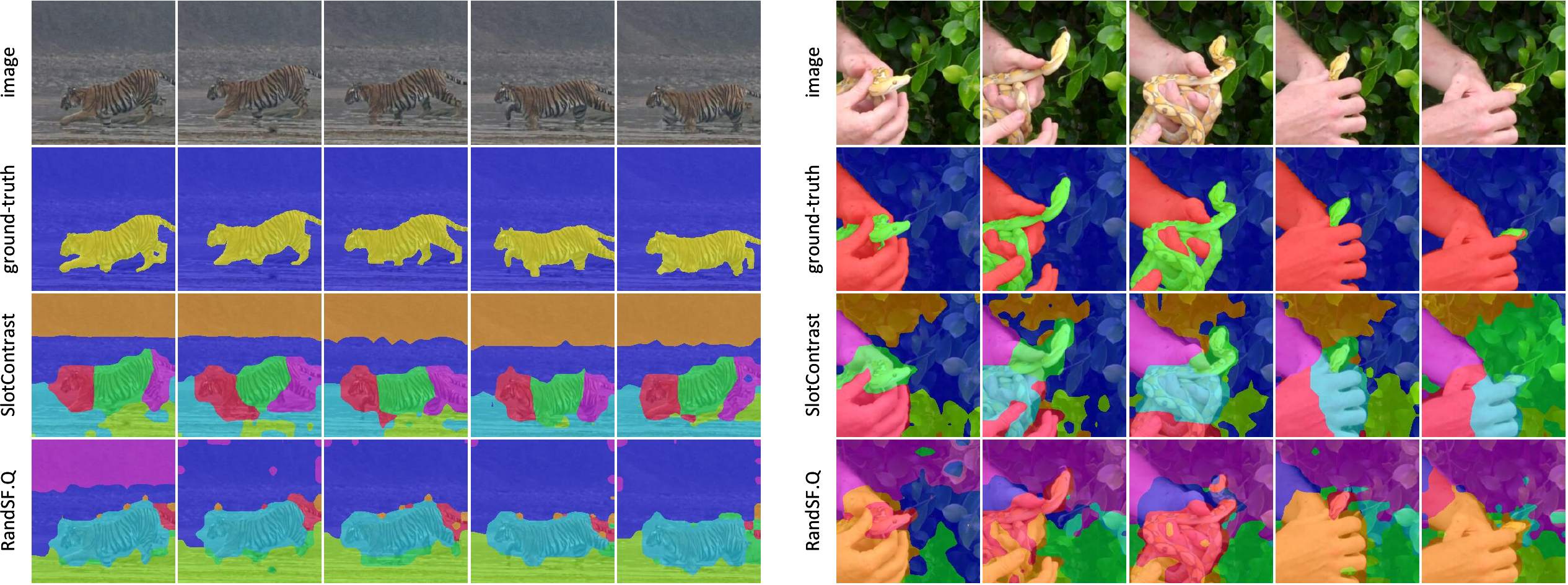}
\caption{
Qualitative results of our RandSF.Q on YTVIS, compared with SotA SlotContrast.
}
\label{fig:qualitative}
\end{figure*}

\subsection{Informative Query Prediction}
\label{sect:query_prediction}

As shown in \Cref{fig:solution} (\textit{right}) and \Cref{fig:teaser} (\textit{b}), we introduce a novel technique, which exploits the latest slot-feature pair for more informative query prediction.

During \textbf{evaluation}, our simplified description of \Cref{eq:transit} is actually expanded with relative time embeddings. Our transitioner $\boldsymbol{\phi}_\mathrm{r}$ takes current slots $\boldsymbol{S}_t$ plus current relative time embedding $\boldsymbol{E}[1] \in \mathbb{R} ^ {1 \times c}$ as the starting point, and conditions on the already-available next feature $\boldsymbol{F}_{t+1}$ plus the next relative time embedding $\boldsymbol{E}[0] \in \mathbb{R} ^ {1 \times c}$ as the latest incremental information, to predict the next query $\boldsymbol{Q}_{t+1}$:
\begin{equation}
\label{eq:transit_eval}
\boldsymbol{\phi}_\mathrm{r} : \boldsymbol{S}_t + \boldsymbol{E}[1], \boldsymbol{F}_{t+1} + \boldsymbol{E}[0] \rightarrow \boldsymbol{Q}_{t+1}
\end{equation}
where $\boldsymbol{E} \in \mathbb{R} ^ {\mathit{\Delta} \times c}$ is a table of learnable relative time embeddings, with window size $\mathit{\Delta} \in [1, T]$, which we will detail in the following subsection; $\cdot[\cdot]$ is indexing operation. 

We assert that next feature $\boldsymbol{F}_{t+1}$ is much more informative than solely current slots $\boldsymbol{S}_{t}$ or all available slots $\{ \boldsymbol{S}_i \} _{i=1}^t$ for predicting next query $\boldsymbol{Q}_{t+1}$. 
This is because, as shown in \Cref{eq:aggregat}, next feature contains every up-to-date information about next slots and of course next query.

\textbf{In contrast}, all existing transitioners predict the next query only based on available slots. 
As shown in \Cref{fig:teaser} (\textit{a}) versus (\textit{b}), none of them utilize next feature.

Specifically, most existing video OCL transitioners $\boldsymbol{\phi}_\mathrm{r}^1$ predict next query $\boldsymbol{Q}_{t+1}$ based only on current slots $\boldsymbol{S}_t$:
\begin{equation}
\label{eq:transit1}
\boldsymbol{\phi}_\mathrm{r}^1 : \boldsymbol{S}_t \rightarrow \boldsymbol{Q}_{t+1}
\end{equation}
where $\boldsymbol{\phi}_\mathrm{r}^1$ is parameterized as a Transformer encoder block \cite{vaswani2017transformer, singh2022steve, kipf2021savi}.
The transitioner $\boldsymbol{\phi}_\mathrm{r}^2$ of remaining video OCL methods predict from all available slots $\{\boldsymbol{S}_i\}_{i=0}^t$:
\begin{equation}
\label{eq:transit2}
\boldsymbol{\phi}_\mathrm{r}^2 : \{ \boldsymbol{S}_i \} _{i=1}^t \rightarrow \boldsymbol{Q}_{t+1}
\end{equation}
where $\boldsymbol{\phi}_\mathrm{r}^2$ is parameterized as some multi-layer Transformer encoder variant \cite{li2025statm, li2025slotpi}.

\textbf{Comment}. Compared with existing transitioner $\boldsymbol{\phi}_\mathrm{r}^1$, our slot-feature pair for informative query prediction only introduces a cross attention submodule and a few time embedding parameters. The extra computation overhead in evaluation is negligible relative to the whole model. But compared with existing transitioner $\boldsymbol{\phi}_\mathrm{r}^2$, our design is multi-fold lightweight as we use only one Transformer block while they use a sequential stack of multiple Transformer blocks.

\begin{table*}[]
\centering\small
\setlength{\aboverulesep}{0pt}  
\setlength{\belowrulesep}{0pt}  
\newcommand{\tss}[1]{\scalebox{0.8}{\texttt{#1}}}
\newcommand{\cg}[1]{\textcolor{green}{#1}}
\newcommand{\std}[1]{\scalebox{0.4}{±#1}}

\begin{tabular}{ccccccccccccc}
\hline
& \multicolumn{4}{c}{MOVi-C {\tiny \#slot=11, conditional}} & \multicolumn{4}{c}{MOVi-D {\tiny \#slot=21, conditional}} & \multicolumn{4}{c}{YTVIS-HQ {\tiny \#slot=7}} \\
\arrayrulecolor{gray}
\cmidrule(lr){2-5} \cmidrule(lr){6-9} \cmidrule(lr){10-13}
\arrayrulecolor{black}
& ARI & ARI\textsubscript{fg} & mBO & mIoU & ARI & ARI\textsubscript{fg} & mBO & mIoU & ARI & ARI\textsubscript{fg} & mBO & mIoU \\
\cmidrule(){1-13}
STEVE      & -- & -- & -- & -- & 32.7\std{0.2} & 66.5\std{0.2} & 23.0\std{0.3} & 21.2\std{0.3} & -- & -- & -- & -- \\
VideoSAUR   & 41.9\std{1.1} & 53.3\std{2.1} & 16.1\std{0.4} & 14.8\std{0.4} & 22.5\std{5.0} & 40.0\std{20.1} & 11.6\std{6.6} & 10.8\std{6.1} & 33.8\std{0.7} & 49.2\std{0.5} & 29.9\std{0.4} & 29.7\std{0.4} \\
\arrayrulecolor{gray}
\cmidrule(){1-13}
\arrayrulecolor{black}
SlotContrast    & 64.6\std{9.4} & 59.9\std{5.3} & 27.7\std{3.0} & 25.8\std{2.9} & 45.3\std{4.1} & 63.9\std{0.2} & 26.7\std{1.0} & 25.1\std{1.0} & 37.2\std{0.6} & 49.4\std{1.1} & 33.0\std{0.2} & 32.8\std{0.1} \\
\cg{RandSF.Q}\textsubscript{\texttt{tsim}}    & 64.0\std{2.9} & \cg{66.3}\std{1.7}	& \cg{28.4}\std{1.3} & \cg{26.1}\std{1.1} & 41.2\std{2.2} & \cg{72.0}\std{1.1} & \cg{27.1}\std{0.9} & \cg{25.4}\std{0.9} & \cg{46.0}\std{0.7} & \cg{60.4}\std{2.3} & \cg{39.4}\std{0.3} & \cg{38.5}\std{0.2} \\  
\cg{RandSF.Q}\textsubscript{\texttt{ssc}}    & \cg{65.4}\std{10.7} & \cg{67.4}\std{2.1}	& \cg{29.2}\std{3.8} & \cg{26.8}\std{3.7} & 41.6\std{3.7} & \cg{77.5}\std{1.0} & \cg{27.4}\std{1.0} & \cg{25.6}\std{1.0} & \cg{40.1}\std{0.4} & \cg{58.0}\std{1.0} & \cg{37.6}\std{0.4} & \cg{37.2}\std{0.4} \\
\hline
\end{tabular}

\\\textcolor{red}{update 20260416 below}\\
\setlength{\aboverulesep}{0pt}  
\setlength{\belowrulesep}{0pt}  

\begin{tabular}{ccccccccccccc}
\hline
& \multicolumn{4}{c}{MOVi-C {\tiny \#slot=11, conditional}} & \multicolumn{4}{c}{MOVi-E {\tiny \#slot=24, conditional}} & \multicolumn{4}{c}{YTVIS-2022 {\tiny \#slot=7}} \\
\arrayrulecolor{gray}
\cmidrule(lr){2-5} \cmidrule(lr){6-9} \cmidrule(lr){10-13}
\arrayrulecolor{black}
& ARI & ARI\textsubscript{fg} & mBO & mIoU & ARI & ARI\textsubscript{fg} & mBO & mIoU & ARI & ARI\textsubscript{fg} & mBO & mIoU \\
\cmidrule(){1-13}
VideoSAUR   & 41.9\std{1.1} & 53.3\std{2.1} & 16.1\std{0.4} & 14.8\std{0.4} & 17.4\std{2.5} & 34.6\std{20.7} & 8.3\std{4.9} & 7.5\std{4.3} & 33.4\std{0.8} & 48.2\std{0.7} & 27.2\std{0.3} & 26.8\std{0.3} \\
\arrayrulecolor{gray}
\cmidrule(){1-13}
\arrayrulecolor{black}
SlotContrast    & 64.6\std{9.4} & 59.9\std{5.3} & 27.7\std{3.0} & 25.8\std{2.9} & 29.9\std{4.9} & 70.6\std{3.8} & 20.7\std{1.4} & 19.3\std{1.2} & 35.2\std{0.8} & 51.4\std{0.7} & 29.7\std{0.5} & 29.3\std{0.6} \\
\cg{RandSF.Q}\textsubscript{\texttt{tsim}}    & 64.0\std{2.9} & \cg{66.3}\std{1.7}	& \cg{28.4}\std{1.3} & \cg{26.1}\std{1.1} & \cg{34.4}\std{2.5} & \cg{74.0}\std{1.3} & \cg{22.9}\std{0.9} & \cg{21.6}\std{0.8} & \cg{40.3}\std{7.0} & \cg{51.6}\std{4.1} & \cg{34.0}\std{3.6} & \cg{33.3}\std{3.7} \\  
\cg{RandSF.Q}\textsubscript{\texttt{ssc}}    & \cg{65.4}\std{10.7} & \cg{67.4}\std{2.1}	& \cg{29.2}\std{3.8} & \cg{26.8}\std{3.7} & \cg{30.5}\std{1.2} & \cg{82.1}\std{3.1} & \cg{23.0}\std{1.2} & \cg{21.6}\std{1.4} & \cg{37.9}\std{1.3} & \cg{51.8}\std{1.2} & \cg{32.2}\std{1.8} & \cg{31.5}\std{1.8} \\  
\hline
\end{tabular}

\caption{
Object discovery on videos.
Input resolution is 256$\times$256 (224$\times$224); DINO2 ViT-S/14 is for encoding.
On YTVIS, RandSF.Q surpasses SotA SlotContrast by 10+ points by ARI and ARI\textsubscript{fg}, and \midtilde7 points by mBO and mIoU.
Note that \texttt{tsim} means using the time similarity loss from VideoSAUR \cite{zadaianchuk2024videosaur}, while \texttt{ssc} means using the slot-slot contrast loss from SlotContrast \cite{manasyan2025slotcontrast}.
}
\label{tab:objdiscov}
\end{table*}

\begin{table}[]
\centering\small
\setlength{\tabcolsep}{3.5pt}
\setlength{\aboverulesep}{0pt}  
\setlength{\belowrulesep}{0pt}  
\newcommand{\tss}[1]{\scalebox{0.8}{\texttt{#1}}}
\newcommand{\cg}[1]{\textcolor{green}{#1}}
\newcommand{\std}[1]{\scalebox{0.4}{±#1}}

\begin{tabular}{c@{}c@{}ccccc}
\hline
\arrayrulecolor{gray}
&&& \multicolumn{4}{c}{YTVIS-HQ {\tiny \#slot=7}} \\
\cmidrule(lr){4-5} \cmidrule(lr){6-6} \cmidrule(lr){7-7}
\arrayrulecolor{black}
&&& class top1$\uparrow$ & top3$\uparrow$ & bbox IoU$\uparrow$ & \#match$\uparrow$ \\
\cmidrule(){1-7}
SlotContrast&+& MLP & 85.8\std{0.3} & 95.8\std{0.4} & 51.5\std{0.5} & 9249\std{41} \\
\cg{RandSF.Q}    &+& MLP & \cg{90.5}\std{0.3} & \cg{97.9}\std{0.3} & 50.6\std{0.4} & 8979\std{123} \\
\arrayrulecolor{black}
\hline
\end{tabular}

\\\textcolor{red}{update 20260416 below}\\
\setlength{\tabcolsep}{3.5pt}
\setlength{\aboverulesep}{0pt}  
\setlength{\belowrulesep}{0pt}  

\begin{tabular}{c@{}c@{}ccccc}
\hline
\arrayrulecolor{gray}
&&& \multicolumn{4}{c}{YTVIS-2022 {\tiny \#slot=7}} \\
\cmidrule(lr){4-5} \cmidrule(lr){6-6} \cmidrule(lr){7-7}
\arrayrulecolor{black}
&&& class top1$\uparrow$ & top3$\uparrow$ & bbox IoU$\uparrow$ & \#match$\uparrow$ \\
\cmidrule(){1-7}
SlotContrast&+& MLP & 87.1\std{0.2} & 96.4\std{0.1} & 48.2\std{0.3} & 19943\std{156} \\
\cg{RandSF.Q}    &+& MLP & \cg{90.7}\std{0.1} & \cg{97.1}\std{0.3} & 46.5\std{0.7} & 19120\std{220} \\
\arrayrulecolor{black}
\hline
\end{tabular}

\caption{
Object recognition on YTVIS.
}
\label{tab:objrecogn}
\end{table}

\subsection{Effective Query Prediction Learning}
\label{sect:query_prediction_learning}

As shown in \Cref{fig:solution} (\textit{right}), we introduce another novel technique, which explores random slot-feature pairs to enforce effective query prediction learning.

During \textbf{training}, our simplified formulation of \Cref{eq:transit} is in fact expanded with random sampling from the available recurrences within a time window. Our transitioner $\boldsymbol{\phi}_\mathrm{r}$ takes slots $\boldsymbol{S}_{t_1}$ at random time step $t_1 \in [t-\mathit{\Delta}+1, t]$ plus the corresponding relative time embedding $\boldsymbol{E}[t+1 - t_1]$, as the starting point, and conditions on feature $\boldsymbol{F}_{t_2}$ at random time step $t_2 \in [t-\mathit{\Delta}+2, t+1]$ plus the corresponding relative time embedding $\boldsymbol{E}[t+1 - t_2]$, as the incremental information, to predict next query $\boldsymbol{Q}_{t+1}$:
\begin{equation}
\label{eq:transit_train}
\boldsymbol{\phi}_\mathrm{r} : \boldsymbol{S}_{t_1} + \boldsymbol{E}[t+1 - t_1], \boldsymbol{F}_{t_2} + \boldsymbol{E}[t+1 - t_2] \rightarrow \boldsymbol{Q}_{t+1}
\end{equation}
where
\begin{equation}
\label{eq:t1}
t_1 \sim \mathcal{U} \{ t - \mathit{\Delta} + 1, ... t \}
\end{equation}
\begin{equation}
\label{eq:t2}
t_2 \sim \mathcal{U} \{ t - \mathit{\Delta} + 2, ... t + 1 \}
\end{equation}
Here $\boldsymbol{E} \in \mathbb{R} ^ {\mathit{\Delta} \times c}$ is the same variable in \Cref{eq:transit_eval}. We empirically set $\mathit{\Delta}$ to be identical to the widely-adopted video clip length in training, e.g., $\mathit{\Delta}=6$ given video length $T=24$ \cite{kipf2021savi,elsayed2022savipp} and $\mathit{\Delta}=5$ given video length $T=20$ \cite{zhao2025dias}.

We deem that a transitioner should be able to predict the query from any starting point slots $\boldsymbol{S}_{t_1}$ that is not too far away, i.e., within a relative small window size $\mathit{\Delta}$, conditioned on the incremental information provided by feature $\boldsymbol{F}_{t_2}$.
Introducing the above-mentioned randomness in training drives the transitioner to grasp such transition dynamics knowledge for better query prediction.

\textbf{By comparison}, existing video OCL transitioners both $\boldsymbol{\phi}_\mathrm{r}^1$ and $\boldsymbol{\phi}_\mathrm{r}^2$ predict the next query naively from current slots, where there is only one time step of difference. 
As shown in \Cref{fig:teaser} (\textit{a}) versus (\textit{c}), such naive design cannot drive the transitioner to effectively grasp the dynamics for query prediction.
Specifically, $\boldsymbol{\phi}_\mathrm{r}^1$ and $\boldsymbol{\phi}_\mathrm{r}^2$ learn query prediction as in \Cref{eq:transit1} and \Cref{eq:transit2}, respectively.

\textbf{Comment}. Same as existing transitioners $\boldsymbol{\phi}_\mathrm{r}^1$ and $\boldsymbol{\phi}_\mathrm{r}^2$, our transitioner $\boldsymbol{\phi}_\mathrm{r}$ is trained inside the OCL model in an end-to-end way, requiring no extra loss. Namely, our random slot-feature pair enables effective query prediction learning while maintaining architectural elegance. Our computation overhead in training is mostly same as that in evaluation.

\section{Experiment}
\label{sect:experiment}

We comprehensively evaluate our method on object discovery, as well as downstream tasks of object recognition and visual question answering.
We also look into how well our method learns transition knowledge for query prediction.
We conduct each experiment using three random seeds whenever it is applicable.

\subsection{Overall Setting}

We cover the following baselines and datasets to ensure comprehensive and fair comparisons.

\textbf{Baselines}.
STEVE \cite{singh2022steve} realizes the first Slot Attention-based OCL on videos.
VideoSAUR \cite{zadaianchuk2024videosaur} is the first method that utilizes vision foundation model and achieves competitive results in OCL on real-world complex videos.
SlotContrast \cite{manasyan2025slotcontrast} enables temporal consistency on long videos, setting new state-or-the-art (SotA) recently.
Our method RandSF.Q is compared with these methods.
Comparing with methods like SOLV \cite{aydemir2023solv}, which uses slot pruning, is unfair.
So does comparing with SAVi \cite{kipf2021savi} and SAVi++ \cite{elsayed2022savipp}, which are trained with external stronger supervision like optical flow and depth map.

\textbf{Datasets}.
We evaluate these methods on both synthetic and real-world video datasets, following the experiments of SotA method SlotContrast.
For synthetic video datasets, we use MOVi-C and MOVi-D of MOVi datasets series\footnote{https://github.com/google-research/kubric/tree/main/challenge\\s/movi}.
These two subsets contain daily objects with incremental complex textures falling on complex background.
These datasets are used for \textit{object discovery}.
For real-world video datasets, we use YouTube Video Instance Segmentation\footnote{https://youtube-vos.org/dataset/vis} (YTVIS) the high quality version\footnote{https://github.com/SysCV/vmt?tab=readme-ov-file\#hq-ytvis-high-quality-video-instance-segmentation-dataset}, which consists of daily-life diverse and complex videos downloaded from YouTube.
This dataset is used for both \textit{object discovery} and \textit{object recognition}.
We also include the synthetic video-text dataset CLEVRER\footnote{http://clevrer.csail.mit.edu}, which consists of geometric objects and paired question texts, testing the multi-modal reasoning capability. This dataset is used for \textit{visual question answering}.

\subsection{Scene Representation}

Video OCL models can be used to represent temporal visual scenes in the object-centric manner.
The representation quality can be evaluated by both object discovery, which is achieved by the binarizing slots' attention maps along the slot dimension into object segmentation masks, and object recognition, which is achieved by predicting objects' class and bounding box from corresponding slots. Note that the former is unsupervised while the latter is supervised.

\textbf{Object discovery}.
We compare our method RandSF.Q with baselines STEVE, VideoSAUR and SlotContrast in terms of their object discovery performance on datasets MOVi-C, MOVi-D and YTVIS.
This is basically an unsupervised object segmentation task.
We use multiple recognized metrics for comprehensive measurement: Adjusted Rand Index (ARI)\footnote{https://scikit-learn.org/stable/modules/generated/sklearn.metri\\cs.adjusted\_rand\_score.html} roughly for background segmentation accuracy measurement, ARI foreground (ARI\textsubscript{fg}) for foreground large objects, mean Best Overlap (mBO) \cite{uijlings2013selectivesearch} for best overlapped regions, and mean Intersection over Union (mIoU)\footnote{https://scikit-learn.org/stable/modules/generated/sklearn.metri\\cs.jaccard\_score.html} as the most strict measurement.

As shown in \Cref{tab:objdiscov}, across all datasets and almost all metrics, our RandSF.Q defeats all the baselines.
Specifically, on dataset YTVIS, RandSF.Q surpasses the latest SotA SlotContrast by more than 10 points by metrics ARI and ARI\textsubscript{fg}, and up to 6 points by mBO and mIoU.
In contrast, SlotContrast only boost their baseline VideoSAUR limitedly, less than 3 points by metrics mBO and mIoU, or even degrading by ARI and ARI\textsubscript{fg}.

\textbf{Object recognition}.
We compare our method RandSF.Q with baseline SlotContrast in terms of their object recognition performance on dataset YTVIS.
This consists of two sub-tasks, i.e., supervised classification and regression from slots.
Some literature \cite{locatello2020slotattent} names this as \textit{set prediction}.
We follow \cite{seitzer2023dinosaur} to represent the dataset as slots and train a two-layer MLP to predict the object class and bounding box corresponding to each slot, supervised by the annotations of object class labels and bounding boxes in the dataset.
We use top1/top3 accuracy and box IoU score as the class label classification and bounding box regression performance metrics.

As shown in \Cref{tab:objrecogn}, our RandSF.Q defeats baseline SlotContrast both in object class label classification and object bounding boxe coordinate regression. 

These two aspects of experiments prove our method's superiority in scene especially object representation.

\begin{table}[]
\centering\small
\setlength{\aboverulesep}{0em}  
\setlength{\belowrulesep}{0em}  
\newcommand{\tss}[1]{\scalebox{0.8}{\texttt{#1}}}
\newcommand{\cg}[1]{\textcolor{green}{#1}}
\newcommand{\std}[1]{\scalebox{0.4}{±#1}}

\begin{tabular}{c@{}c@{}ccc}
\hline
\arrayrulecolor{gray}
&&& \multicolumn{2}{c}{CLEVRER {\tiny \#slot=7}} \\
\cmidrule(lr){4-5}
\arrayrulecolor{black}
&&& per option \% & per question \% \\
\cmidrule(){1-5}
SlotContrast&+& Aloe & 97.2\std{1.1} & 95.6\std{0.9} \\
\cg{RandSF.Q}&+& Aloe & \cg{98.5}\std{0.8} & \cg{96.3}\std{0.7} \\
\arrayrulecolor{black}
\hline
\end{tabular}

\caption{
Visual question answering on CLEVRER.
}
\label{tab:vqa_on_clevrer}
\end{table}


\subsection{Scene Understanding}

Better object representation provides more information for understand and reasoning about the visual scene.

\textbf{Visual question answering}.
We compare our method RandSF.Q with baseline SlotContrast in terms of their visual question answering performance on dataset CLEVRER.
We follow the data processing and model design of Aloe \cite{ding2021aloe, wu2022slotformer}. 
We firstly pre-train OCL models on CLEVRER and then freeze them.
Each video's frames are all represented as slots, then slots of different frames are added with corresponding time step embeddings; meanwhile, question texts are embedded into vectors and added with corresponding position embeddings.
These tokens are fed into Aloe together, appended with a classification token.
The output is obtained by projecting the transformed classification token into logits of all answer labels.

As shown in \Cref{tab:vqa_on_clevrer}, our method RandSF.Q defeats baseline SlotContrast in performance measured either by per option accuracy or by per question accuracy.
As objects in CLEVRER's synthetic videos are too simple and the baseline performance is already high enough, our method's performance boosts are relatively less significant, compared with those in object discovery and recognition.

\subsection{Ablation}

\begin{table}[]
\centering\small
\setlength{\tabcolsep}{4pt}
\setlength{\aboverulesep}{0pt}  
\setlength{\belowrulesep}{0pt}  
\newcommand{\tss}[1]{\scalebox{0.8}{\texttt{#1}}}
\newcommand{\cg}[1]{\textcolor{green}{#1}}
\newcommand{\std}[1]{\scalebox{0.4}{±#1}}

\begin{tabular}{ccccc}
\hline
utilizing next feature  & \checkmark    &   & \checkmark    & \checkmark \\
sampling slot-feature pair  & \checkmark    & \checkmark    &   & \checkmark \\
injecting relative time & \checkmark    & \checkmark    & \checkmark    &   \\

\arrayrulecolor{gray}
\cmidrule(lr){1-1}\cmidrule(lr){2-5}
\arrayrulecolor{black}

ARI+ARI\textsubscript{fg} & 108.0\std{1.3} & 99.7\std{3.2} & 81.6\std{11.4} & 64.6\std{4.7}\\
\hline

&&&& \\

\hline
sampling window size & 2 & 3 & 4 & 5 \\

\arrayrulecolor{gray}
\cmidrule(lr){1-1}\cmidrule(lr){2-5}
\arrayrulecolor{black}

ARI+ARI\textsubscript{fg}   & 90.0\std{3.3} & 102.0\std{4.1} & 107.8\std{0.4} & 108.0\std{1.3} \\
\hline

&&&& \\

\hline
time injecting method & \multicolumn{2}{c}{append time emb} & \multicolumn{2}{c}{sum time emb} \\

\arrayrulecolor{gray}
\cmidrule(lr){1-1}\cmidrule(lr){2-5}
\arrayrulecolor{black}

ARI+ARIfg             & \multicolumn{2}{c}{98.8\std{3.9}}   & \multicolumn{2}{c}{108.0\std{1.3}} \\
\hline
\end{tabular}

\begin{tabular}{cc} & \end{tabular}

\caption{
Ablation study.
}
\label{tab:ablat}
\end{table}

As shown in \Cref{tab:ablat}, we ablate the effects of our model designs from the following perspectives.

\textbf{Utilizing next feature}: Whether to utilize next feature for query prediction?
Utilizing next feature as condition, along with the input slots as starting point for our transitioner does provide more information for query prediction.

\textbf{Random sampling slot-feature pair}: Whether to randomly sample slot-feature pairs for query prediction learning?
Randomly sampling slot-feature pairs from available recurrences ensures better learning of query prediction.

\textbf{Window size $\mathit{\Delta}$}: If sampling, then what window size to use? 2, 3, 4 or 5?
Relatively larger window size is better but the performance saturates when window size is equal to the typical video clip size. By the way, window size larger than the video clip size is not practically possible.

\textbf{Injecting relative time}: Whether to inject relative time into our transitioner?
Injecting relative time of slots and features respectively is very important to tell the transitioner where is the starting point and where is the incremental information for query prediction.

\textbf{Injection method}: If injecting, then how to inject relative time? Appending the relative time embeddings to slots and features respectively or broadcast summing them to slots and features respectively?
Broadcasting and summing time embeddings is obviously better than appending.

\section{Discussion}

Let us go back to our core claim:
\textbf{Does our transitioner truly grasp the knowledge of transition dynamics for query prediction}?

For analysis, we use non-latest slot-feature pairs for query prediction in evaluation, then we count the video OCL performance.
Intuitively, non-latest slot-feature pairs contain less up-to-date information thus leading to inferior query prediction and ultimately inferior video OCL performance.
If our transitioner really learns transition dynamics, then it would still predict good enough queries in such cases.

Note that we do not visualize ARI, ARI\textsubscript{fg}, mBO and mIoU separately. 
Their values often only reflect some specific aspects of video OCL performance, instead of the overall performance, and also often lack clear regularity.
Also note that it is not straightforward to evaluate our query prediction by calculating some distance between the query predicted from non-latest and latest slot-feature pairs, compared with the indirect evaluation scheme described above.
In addition, we ensure the color bar is within identical value range sizes for easy comparison.
We also put the overall performance of baseline SlotContrast here for simple comparison.

\begin{figure}
\centering
\includegraphics[width=\linewidth]{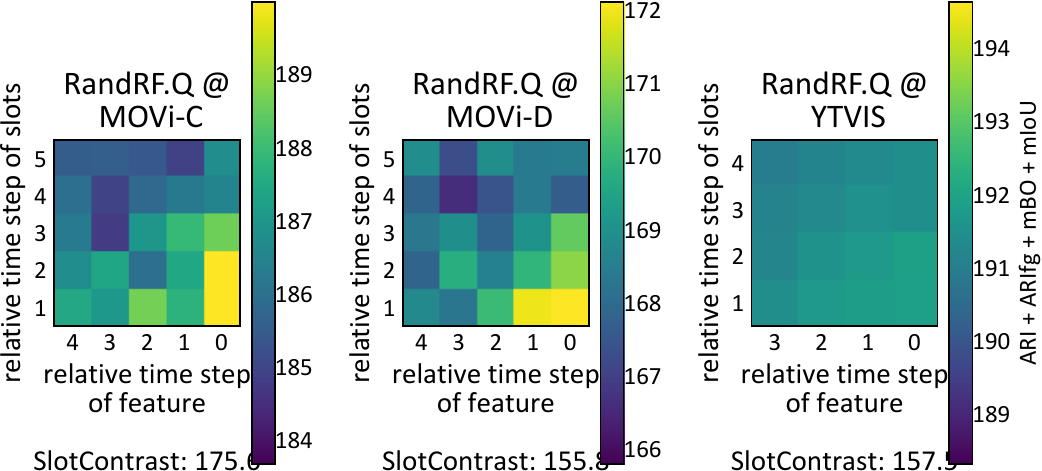}
\caption{
RandSF.Q performance with queries predicted from slot-feature pairs at different relative time steps.
}
\label{fig:acc_matrix_randsfq}
\end{figure}

As shown in \Cref{fig:acc_matrix_randsfq}, \textbf{the elements of each matrix are roughly brighter on the bottom-right}, compared with those on the top-left.
This aligns with our intuition that the more up-to-date the slots and feature being used for the transitioner to predict query, the better the query is and the better our RandSF.Q model performs.
Although using non-latest slot-feature pairs for query prediction degrades the performance, \textbf{our RandSF.Q still always has significant performance advantage over SlotContrast} even when using the oldest slot-feature pair for query prediction.

Interestingly, the performance matrix of RandSF.Q on dataset YTVIS shows little performance fluctuation compared with those on MOVi-C and D.
We explain this as foreground objects in human-shot videos are usually well tracked. 
This means queries from the near past differ little and thus the query prediction is relatively easier.
In contrast, objects in MOVi-C and D drop and bounce chaotically, not as well centered as those in human-shot videos.

\section{Conclusion}
\label{sect:conclusion}

Based on our two observations, we propose our method RandSF.Q. 
We realize informative query prediction by utilizing the next feature and effective query prediction learning by randomly sampling slot-feature pairs from available recurrences as transitioner inputs for query prediction.
Our method show significant performance advantages over the latest SotA methods on scene representation and scene understanding.
Our core claim is also verified through performance matrix analysis.
Our work pushes the video OCL research greatly forward.

\textbf{Limitations and future work}. 
Our method still cannot solve the long-existing issue of the number of slots often mismatches with the real number of objects in a visual scene.
And if introducing those adaptive slot number techniques, our random slot-feature pair sampling technique cannot be directly applied. Sophisticated designs are necessary to combine these two types of techniques.

\section{Acknowledgments}

We acknowledge the support of Finnish Center for Artificial Intelligence (FCAI), Research Council of Finland flagship program.
We thank the Research Council of Finland for funding the projects ADEREHA (grant no. 353198), BERMUDA (362407) and PROFI7 (352788).
We also appreciate CSC - IT Center for Science, Finland, for granting access to supercomputers Mahti and Puhti, as well as LUMI, owned by the European High Performance Computing Joint Undertaking (EuroHPC JU) and hosted by CSC Finland in collaboration with the LUMI consortium.
Furthermore, we acknowledge the computational resources provided by the Aalto Science-IT project through the Triton cluster.

\bibliography{aaai2026}

\end{document}